\documentclass[10pt,twocolumn,letterpaper]{article}

\usepackage[pagenumbers]{cvpr} 

\usepackage{graphicx}
\usepackage{amsmath}
\usepackage{amssymb}
\usepackage{booktabs}
\usepackage{mathtools}
\newtheorem{theorem}{Theorem}

\usepackage[pagebackref,breaklinks,colorlinks]{hyperref}
\usepackage{xspace}
\usepackage{multirow}
\usepackage{algorithm}  
\usepackage{algpseudocode}

\makeatletter
\DeclareRobustCommand\onedot{\futurelet\@let@token\@onedot}

\def\@onedot{\ifx\@let@token.\else.\null\fi\xspace}

\def\eg{\emph{e.g}\onedot} 
\def\ie{\emph{i.e}\onedot} 
 
 \def\vs{\emph{vs}\onedot}
\def\wrt{w.r.t\onedot} 
\def\iid{\emph{i.i.d}\onedot}
\makeatother

\usepackage[capitalize]{cleveref}
\crefname{section}{Sec.}{Secs.}
\Crefname{section}{Section}{Sections}
\Crefname{table}{Table}{Tables}
\crefname{table}{Tab.}{Tabs.}

\def\cvprPaperID{10073} 
\def\confName{CVPR}
\def\confYear{2024}

\begin{document}

\title{Optimizing Few-step Sampler for Diffusion Probabilistic Model}

\author{Jen-Yuan, Huang\\
Peking University\\
{\tt\small jenyuan@stu.pku.edu.cn}
}
\maketitle

\begin{abstract}
   Diffusion Probabilistic Models (DPMs) have demonstrated exceptional capability of generating high-quality and diverse images, but their practical application is hindered by the intensive computational cost during inference. The DPM generation process requires solving a Probability-Flow Ordinary Differential Equation (PF-ODE), which involves discretizing the integration domain into intervals for numerical approximation. This corresponds to the sampling schedule of a diffusion ODE solver, and we notice the solution from a first-order solver can be expressed as a convex combination of model outputs at all scheduled time-steps. We derive an upper bound for the discretization error of the sampling schedule, which can be efficiently optimized with Monte-Carlo estimation. Building on these theoretical results, we purpose a two-phase alternating optimization algorithm. In Phase-1, the sampling schedule is optimized for the pre-trained DPM; in Phase-2, the DPM further tuned on the selected time-steps. Experiments on a pre-trained DPM for ImageNet64 dataset demonstrate the purposed method consistently improves the baseline across various number of sampling steps.
\end{abstract}

\section{Introduction}
\label{introduction}
The great advancement in deep generative model has opened up new possibilities to both Machine Learning community and our living-world. In Computer Vision, this trend is led by Diffusion Probabilistic Model (DPM)~\cite{sohl2015deep, ho2020denoising, song2020score} which presents impressive creativity that originally believed to be only possessed by mankind. DPM does not rely on a surrogate loss neither an adversarial training, making it efficient to train and scale easily with data magnitude and model size. Breaking records left by other generative models on various tasks including unconditional~\cite{ho2020denoising} and conditional image generation, likelihood estimation~\cite{song2021maximum, kingma2021variational}, inpainting~\cite{rombach2022high, ho2020denoising} and super-resolution~\cite{wang2023exploiting}.

DPM was initially inspired by molecular dynamic system~\cite{sohl2015deep} and introduced two stochastic processes to describe the gradual changes in image over time. In the forward process, images are corrupted by progressively added Gaussian noises. Whereas in the reverse process, information is reconstructed from noise. DPM is trained to denoise images corrupted with different amount of noise called \textit{noise levels}. The reverse generation process is then a step-by-step inference on less noisy inputs. Song~\etal~\cite{song2020score} proposed a generalized class of DPM in continuous cases. They revealed the generation process is solving a Stochastic Differential Equation. Though in practice we still inference on the discrete basis, one can regard the specification of noise levels as step-sizes of the numerical solver in continuous DPM. Simulating this process is extremely expensive as thousands of iterations with model called are required to achieve lower global error ,\ie higher fidelity. This property of DPM imposes a natural trade-off between generation quality and speed, preventing a wider application where computational resources are often restricted.

Previous work~\cite{nichol2021improved, song2020denoising, karras2022elucidating} found an evenly distributed noise levels sub-optimal. It is beneficial to take smaller step-sizes when the noise level is low, and steps with relatively high noise level can be skipped without risks of losing too much quality. This empirical discovery has long been adopted by practitioners and intrigued explorations towards sampling algorithms for DPM~\cite{song2020denoising}. Higher-order solvers~\cite{karras2022elucidating} or adaptive solvers~\cite{jolicoeur2021gotta} from the literature of numerical solving differential equations have been attempted. However, these methods are proposed without awareness of DPM and do not fully utilize its properties. Some other methods~\cite{ho2020denoising,watson2021learning} manipulate the underlying probabilistic model of DPM and build a deterministic generation process that shares the same marginal distribution as their stochastic counterparts. In this way, they reduce the number of iterations required by diffusion sampling to a large extent, but the noise level used at each step, \ie, step-size of the solver is chosen empirically. Besides, both of these methods say nothing about truncation error, the optimization of DPM generation process still lacks of theoretical basis.

In this study, we focus on the optimization of DPM as a numerical differential equation solver. We analyze the truncation error of common deterministic DPM sampler~\cite{karras2022elucidating}, and discover the deep connection between sampling schedule and DPM training objective. Finally, we derive a two-stage algorithm based on these analysis for optimizing DPM from the practical perspective. Our contributions are as follows:
\begin{enumerate}
    \item We present theoretical analysis in the truncation error of common first-order DPM sampling algorithm;
    \item We propose an efficient method to optimize diffusion generation process with some per-sample loss;
    \item Based on these analysis, we derive a novel two-stage method to finetune pretrained models as a numerical differential equation solver for a given sampling steps.
\end{enumerate}
\section{Preliminaries}
\label{background}

\subsection{Diffusion Probabilistic Models}

Ho~\etal~\cite{ho2020denoising} defined DPM with two stochastic processes named forward and reverse process. For an image data $\boldsymbol{x}$, the forward process progressively adds small amount of \iid Gaussian noise to it. Due to the Markovian property, the distribution of such a noisy version of image $\boldsymbol{x}$ is:
\begin{equation}
\label{anc_sampling}
    \boldsymbol{x}_i\sim\mathcal{N}\left( \sqrt{1-\sigma_i}\boldsymbol{x}, \sqrt{\sigma_i}\boldsymbol{I} \right),
\end{equation}
where $\sigma_i$ is called the noise level at step $i$. As our primary object is to analyze the reverse process, we denote the minimum noise level as $\sigma_T$ and the maximum as $\sigma_0$. One can use re-parameterization of Eq.~\ref{anc_sampling} and easily sample $\boldsymbol{x}_i$ with:
\begin{equation}
\label{reparam}
    \boldsymbol{x}_i=\boldsymbol{x}+\sigma_i\boldsymbol{\epsilon},
\end{equation}
where $\boldsymbol{\epsilon}$ is a standard Gaussian random variable with the same shape as $\boldsymbol{x}$. Information from $\boldsymbol{x}$ eventually vanish as $i$ goes to zero, and $\boldsymbol{x}_0$ becomes in-distinguishable with pure noise. In the reverse process a neural network is applied to reconstruct images from noise. The magical thing in DPM is that this difficult procedure only requires a model learned to denoise a noisy version of $\boldsymbol{x}$ with different noise level $\sigma_i$, and the training objective of DPM can be written as:
\begin{equation}
\label{diffusion_loss}
    \mathcal{L}_{diff}=\mathbb{E}_{\boldsymbol{x},\boldsymbol{n},i}\left[\lambda_i\lVert D_{\theta}\left(\boldsymbol{x}_i,\sigma_{i}\right)-\boldsymbol{x} \rVert^2\right],
\end{equation}
where the denoising function $D_\theta$ explicitly conditions on noise level $\sigma_i$. $\lambda_i$ is a weighting factor that is often adopted to enhance training on specific noise levels.

\subsection{Diffusion Differential Equation}
\label{DDE}
Song~\etal~\cite{song2020score} generalizes the above Markov-based DPM into continuous cases where the two processes are expressed by Stochastic Differential Equation (SDE). Specifically, for the forward process we have an SDE:
\begin{equation}
\label{fwd-sde}
d\boldsymbol{x}=f(\boldsymbol{x},t)dt+g(t)d\boldsymbol{w},
\end{equation}
where $f(\boldsymbol{x},t)$ is the drift coefficient and $g(t)$ the diffusion coefficient. The noise level is now a continuous function $\sigma(t)$ for $t\in\left[0,1\right]$. The marginal distribution $p(\boldsymbol{x};t)$ induced by this SDE is still a Gaussian $\mathcal{N}\left( s(t)\boldsymbol{x}, s^2 (t)\sigma^2 (t)\right)$ where $s(t)$ is derived from drift coefficient $f(\boldsymbol{x},t)$~\cite{karras2022elucidating}. Correspondingly, we have another SDE:
\begin{equation}
\label{rvs-sde}
    d\boldsymbol{x}=\left(f(\boldsymbol{x},t)-g^2(t)\nabla_{\boldsymbol{x}}\log p(\boldsymbol{x};t)\right)dt+g(t)d\Bar{\boldsymbol{w}},
\end{equation}
for the reverse process. $\boldsymbol{w}$ and $\Bar{\boldsymbol{w}}$ are standard Weiner process and its reverse. Song~\etal also introduced Probability-Flow Ordinary Differential Equation (PF-ODE) which is a deterministic process once the initial point is set, but shares the same marginal distribution $p(\boldsymbol{x};t)$ as Eq.~\ref{rvs-sde}. Comparing to its stochastic counterpart, PF-ODE is much more stable in simulating and provides possibility in large-step integration.

The choice of $f(\boldsymbol{x},t)$ and $g(t)$ depend on parameterizations of the underlying diffusion process. Variance Preserving (VP) and Variance Exploding (VE) are two most widely used instances~\cite{song2020score} with $\sigma_0=1$ and $\sigma_{max}$ accordingly. Karras~\etal~\cite{karras2022elucidating} demonstrate both VP and VE models are variations to a canonical one with PF-ODE:
\begin{equation}
  d\boldsymbol{x}=-\sigma(t)\nabla_{\boldsymbol{x}}\log p(\boldsymbol{x};\sigma(t))d\sigma(t).
  \label{ODE}
\end{equation}
Our analysis will be conducted on this canonical ODE, and we will later show in \S\ref{improve} that our method is valid for both different modeling with consistent improvements.

\subsection{Sampling Schedule}

The SDE formulation in \S\ref{DDE} is in the context of continuous diffusion process and is only for theoretical analysis purpose. In practice $\sigma(t)$ must be discretized into a strictly decreasing sequence of noise levels $\sigma_i,i\in\left\{0,...,T\right\}$ called sampling schedule. The nabla term in Eq.~\ref{ODE} is modeled by denoising function that $\nabla_{\boldsymbol{x}}\log p(\boldsymbol{x};\sigma(t))=\frac{D_{\theta}(\boldsymbol{x}_{i}, \sigma_i)-\boldsymbol{x}_{i}}{\sigma_{i}^{2}}$. Most previous works treat sampling schedule as a hyper-parameter[]. However numerical solving Eq.~\ref{ODE} will inevitably induce truncation error, the empirically chosen step-sizes are by no means optimal and may vary across different models. Given the major drawback of DPM, identify the optimal sampling schedule for numerically solving Eq.~\ref{ODE}, and the denoising model $D_\theta$ optimized for this sampling schedule can jointly reduce the truncation error of DPM as a numerical differential equations solver, allowing for larger step-sizes and fewer iterations to alleviate the limitations of DPM.
\section{Method}
\label{method}

\subsection{Discretization Loss}
\label{disc_loss}
\begin{figure*}
  \centering
    \includegraphics[width=1.0\linewidth]{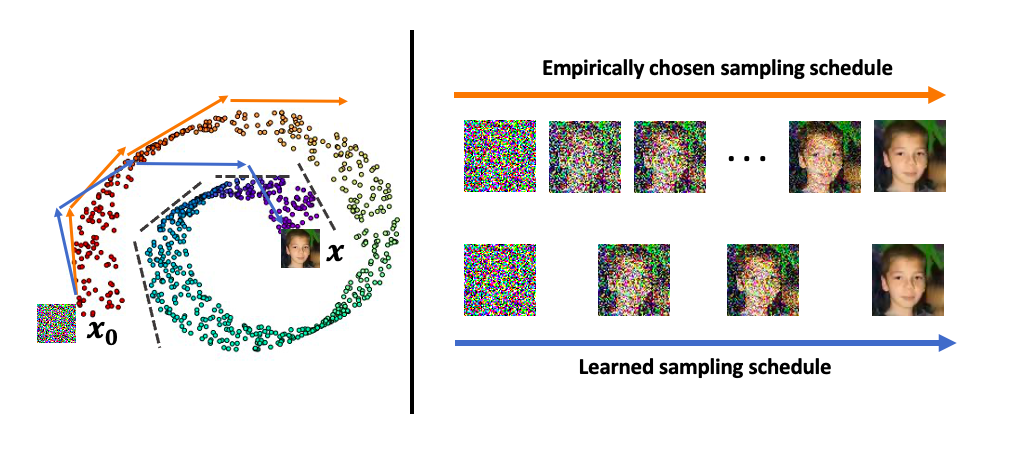}
  \caption{Illustration of the optimal sampling schedule (blue line). The iteration process is following backward Euler-steps, leading to faster converge with low global-error. While the fixed sampling schedule is either sub-optimal, or take equal step-sizes in sampling.}
  \label{fig:illus}
\end{figure*}
To analyze the truncation error of DPM reverse process, consider the most commonly adopted Euler method~\cite{karras2022elucidating,song2020score}. An Euler-step of numerically solving Eq.~\ref{ODE} at step $i$ is:
\begin{equation}
\begin{split}
\label{iteration}
\boldsymbol{\hat{x}}_{t-1} &= \boldsymbol{\hat{x}}_t + \left( \sigma_{t-1}-\sigma_{t} \right)\frac{\boldsymbol{\hat{x}}_{t}-D_{\theta} \left( \boldsymbol{\hat{x}}_{t},\sigma_t \right)}{\sigma_{i}}\\
&=\frac{\sigma_{t-1}}{\sigma_t}\boldsymbol{\hat{x}}_{t}+\left( 1-\frac{\sigma_{t-1}}{\sigma_t} \right)D_{\theta} \left( \boldsymbol{\hat{x}}_{t},\sigma_t \right),
\end{split}
\end{equation}
where the iteration sequence is denoted as $\left\{\boldsymbol{\hat{x}}_i,i=0,...T\right\}$. In the second equation of Eq.~\ref{iteration} we rewrite the forward Euler-step as a linear combination of current iteration point $\boldsymbol{\hat{x}}_i$ and its denoise output $D_{\theta} \left( \boldsymbol{\hat{x}}_{i},\sigma_i \right)$ with factor $\sigma_{i+1}/\sigma_{i}$. For a strictly decreasing noise level these factors lie within interval $\left(0,1\right)$, and Eq.~\ref{iteration} elucidates the reconstruction of DPM reverse process is by continuously adjusting generated contents through the incorporation of new denoise predictions at each step. Eq.~\ref{iteration} can be recursively applied to sequence $\left\{\boldsymbol{\hat{x}}_i\right\}$ and $\boldsymbol{\hat{x}}_{T}$ can then be expanded into a weighted summation of denoise outputs at each iteration:
\begin{equation}
\begin{split}
\label{expansion}
\boldsymbol{\hat{x}}_{0}=\sum_{t=T}^{1}\lambda_{t}D_{\theta} \left( \boldsymbol{\hat{x}}_t,\sigma_t \right)+\frac{\sigma_1}{\sigma_T}\boldsymbol{x}_T, \lambda_{t}=\frac{\sigma_{1}}{\sigma_{t-1}}-\frac{\sigma_{1}}{\sigma_t}
\end{split}
\end{equation}
where $\lambda_{i}=\frac{\sigma_{T}}{\sigma_{i+1}}-\frac{\sigma_{T}}{\sigma_i}$. Weights in Eq.~\ref{expansion} capture the rate of change in the reciprocal of noise levels that denoise outputs with lower input noise levels are granted higher weights in $\boldsymbol{\hat{x}}_T$. This interpretation aligns with observations of previous works~\cite{dhariwal2021diffusion,choi2022perception}; they visualized diffusion generation process and found little perceptual contents are generated when it is too noisy for the network to reconstruct any information.

Our second key observation is the final output $\boldsymbol{\hat{x}}$ of iterating Eq.~\ref{iteration} is solely given by $D_{\theta} \left( \boldsymbol{\hat{x}}_{T},\sigma_{T} \right)$ due to $\sigma_{T+1}=0$. From this perspective, sample quality only depends on the capability of denoise model $D_\theta\left(\boldsymbol{\cdot}~,\sigma_{T}\right)$ and input $\boldsymbol{\hat{x}}_{T}$. The former will not be the bottleneck because the noise level is quite small at this point, leaving room for improvements to the latter. An iteration point that is well converged to the data manifold can lead to potential boost in sample quality. Consider the L2-error of at iteration step $T$:
\begin{equation}
\label{T-error}
    \begin{split}
        \mathcal{E}_{T}=\left\lVert \boldsymbol{\hat{x}}_1-\boldsymbol{x}_1 \right\rVert^{2},
    \end{split}
\end{equation}
where $\boldsymbol{x}_T$ is sampled with and $\boldsymbol{\epsilon}_T$ using Eq.~\ref{reparam}. Given a fixed denoising model $D_\theta$, Eq.~\ref{T-error} is then a direct consequence of the sampling schedule.By bringing in Eq.~\ref{expansion} we propose the discretization loss \wrt a sampling schedule:
\begin{equation}
\label{loss}
    \begin{split}
        \mathcal{L}_{disc}\left( \boldsymbol{\sigma} \right)=\mathbb{E}_{\boldsymbol{x,\epsilon_T}}\left\lVert \sum_{t=T}^{1}\lambda_{t} D_{\theta} \left( \boldsymbol{\hat{x}}_t,\sigma_t \right)+\frac{\sigma_1}{\sigma_T}\boldsymbol{x}_0-\boldsymbol{x}_1 \right\rVert^{2},
    \end{split}
\end{equation}
where we denote $\left\{\sigma_i,i=0,...,T\right\}$ as $\boldsymbol{\sigma}$. We next demonstrate sampling according to an optimal schedule $\boldsymbol{\sigma^*}$ minimizing Eq.~\ref{loss} provides a lower global-error, \ie, $\left\lVert \boldsymbol{\hat{x}} - \boldsymbol{x} \right\rVert^{2}$.

Consider the sequence of denoise outputs $\left\{D_\theta\left(\boldsymbol{\hat{x}}_i,\sigma_i\right)\right\}$. Given the initial point $\boldsymbol{x}_0=\boldsymbol{x}+\sigma_0\boldsymbol{\epsilon}_0$, $\left\{\boldsymbol{\hat{x}}_i\right\}$ is then a deterministic process and so do their denoise outputs. We denote such a sequence of $\left\{D_\theta\left(\boldsymbol{\hat{x}}_i,\sigma_i\right)\right\}$ as $\left\{\boldsymbol{\tau}_{\sigma_i}\left(\boldsymbol{x}_{0}\right)\right\}$, and function $\boldsymbol{\tau}\left(\sigma\left(t\right);\boldsymbol{x}_0\right)=D_{\theta}\left(\boldsymbol{x}\left(t\right);\sigma\left(t\right)\right)$ where $\boldsymbol{x}\left(t\right)$ is an integral of Eq.~\ref{ODE} from $0$ to $t$ for $t\in[0,1]$. As $\sigma\left(t\right)$ is a monotonic function, we can consider this ground-truth trajectory in $1/\sigma\left(t\right)$ space with change-of-variable. By definition we have $\boldsymbol{\tau}\left(1/\sigma\left(1\right);\boldsymbol{x}_0\right)=\boldsymbol{x}$, and the remaining demonstration can be done by showing the end point of trajectory $\left\{\boldsymbol{\tau}_{\sigma^*_i}\left(\boldsymbol{x}_{0}\right)\right\}$ converges to $\boldsymbol{\tau}\left(1/\sigma\left(1\right);\boldsymbol{x}_0\right)$.
\begin{theorem}
\label{theorem}
For $\boldsymbol{\sigma^*}=\mathop{\arg\min}_{\boldsymbol{\sigma}}\mathcal{L}_{disc}\left( \boldsymbol{\sigma} \right)$, the backward difference of the $i$-th element in sequence $\left\{\boldsymbol{\tau}_{\sigma^*_i}\left(\boldsymbol{x}_{0}\right)\right\}$ equals to the first-order increment in $\boldsymbol{\tau}\left(\frac{1}{\sigma};\boldsymbol{x}_0\right)$ with step-size $\frac{1}{\sigma^*_{i+1}}-\frac{1}{\sigma^*_{i}}$ for $i=1,...,T-1$.
\end{theorem}
Theo.~\ref{theorem} is obtained by setting the derivative of Eq.~\ref{loss} \wrt each $\sigma_i$ to zero. Detailed proof can be found in Appendix. Theo.~\ref{theorem} shows that in the optimal sampling schedule, the sequence of denoise outputs $\left\{D_\theta\left(\boldsymbol{\hat{x}}_i,\sigma^*_i\right)\right\}$ approximates function $D_{\theta}\left(\boldsymbol{x}\left(t\right);\sigma\left(t\right)\right)$ with backward Euler-steps. As its end point $D_\theta\left(\boldsymbol{\hat{x}}_T,\sigma^*_T\right)$ is also the sample output $\boldsymbol{\hat{x}}$, the optimal sampling schedule thus guarantees a first-order global error in both $\left\{D_\theta\left(\boldsymbol{\hat{x}}_i,\sigma_i\right)\right\}$ and $\left\{\boldsymbol{\hat{x}}_i\right\}$, which could not be found in an arbitrarily chosen sampling schedule. We provide a intuitive illustration in Fig.~\ref{fig:illus}.

\begin{algorithm}[h]
  \caption{Two-stage Finetune}  
  \label{alg}  
  \begin{algorithmic}[1]
    \Require  
        Data set $\mathcal{D}$; Denoising model $D_\theta$; Sampling steps ${T}$; Loss scaler $\gamma$; Stage-1 length $N_1$; Stage-2 length $N_2$
    \State Initialize sampling schedule $\boldsymbol{\sigma}$
    \While{not converged} 
        \State Sample batch data $\boldsymbol{x}\sim\mathcal{D}$
        \While{in Stage-1}
            \State Sample $\boldsymbol{\epsilon}\sim\mathcal{N}\left(\boldsymbol{0},\boldsymbol{I}\right)$
            \State $\mathcal{L}_{disc}=\left\lVert D_\theta\left(\boldsymbol{x}+\sigma_T\boldsymbol{\epsilon},\sigma_T\right)-\boldsymbol{x}\right\rVert^2$
            \For{$t=0$ to $T$}
                \State Sample $\epsilon\sim\mathcal{N}\left(\boldsymbol{0},\boldsymbol{I}\right)$
                \State Get weights $\lambda_t$ according to $\boldsymbol{\sigma}$
                \State $\mathcal{L}_{diff}=\lambda_t\left\lVert D_\theta\left(\boldsymbol{x}+\sigma_t\boldsymbol{\epsilon},\sigma_t\right)-\boldsymbol{x}\right\rVert^2$
                \State Estimate $\nabla_{\sigma}\mathcal{L}_{disc}$
                \State Update $\boldsymbol{\sigma}$ with $\nabla_{\sigma}\mathcal{L}_{diff}+\gamma\nabla_{\sigma}\mathcal{L}_{disc}$
            \EndFor
		\EndWhile   
        \While {in Stage-2}
            \State Sample $t\sim\left\{0,...,T\right\}$
            \State Sample $\boldsymbol{\epsilon}\sim\mathcal{N}\left(\boldsymbol{0},\boldsymbol{I}\right)$
            \State Get weights $\lambda_t$ according to $\boldsymbol{\sigma}$
            \State $\mathcal{L}_{diff}=\lambda_t\left\lVert D_\theta\left(\boldsymbol{x}+\sigma_t\boldsymbol{\epsilon},\sigma_t\right)-\boldsymbol{x}\right\rVert^2$
            \State Update $\theta$ with $\nabla_{\theta}\mathcal{L}_{diff}$
		\EndWhile
  	\EndWhile
  \end{algorithmic}  
\end{algorithm}

\subsection{Two-stage Finetune}
In Eq.~\ref{loss} we represent the discretization loss as a function of weighted denoise outputs. By extracting $D_\theta\left(\boldsymbol{x}_0,\sigma_0\right)$ from the summation Eq.~\ref{loss} can be further simplified as:
\begin{equation}
\label{preUB}
    \begin{split}
        \mathcal{L}_{disc}\left( \boldsymbol{\sigma}\right) &=\left\lVert \sum_{t=T}^{1}\lambda_{t}\left(D_{\theta} \left( \boldsymbol{\hat{x}}_t,\sigma_t\right) -\boldsymbol{x}\right)+\sigma_1\left(\boldsymbol{\epsilon}_T-\boldsymbol{\epsilon}_1\right) \right\rVert^{2} \\
        &\leq \left\lVert \sum_{t=T}^{1}\lambda_{t}\left(D_{\theta} \left( \boldsymbol{\hat{x}}_t,\sigma_t \right) -\boldsymbol{x}\right) \right\rVert^{2} +\sigma_{1}^2\left\lVert \boldsymbol{\epsilon}_T-\boldsymbol{\epsilon}_1 \right\rVert^{2},
    \end{split}
\end{equation}
where we re-parameterize $\boldsymbol{x}_T$ and $\boldsymbol{x}_0$, and define $\lambda_0=\frac{\sigma_{T}}{\sigma_{1}}$ for notation simplicity. Note that $\boldsymbol{\epsilon}$ are \iid Gaussian noise by definition so we have $\sigma_{T}^2\left\lVert \boldsymbol{\epsilon}_0-\boldsymbol{\epsilon}_T \right\rVert^{2}=2\sigma^2_T$. As for the first term, it is easy to verify $\sum_{i=0}^{T-1}\lambda_i=1$. Recall $\lambda$ are in the range of $\left(0,1\right)$, the first term in Eq.~\ref{preUB} is actually a square of convex combination and is upper-bounded by:
\begin{equation}
\label{UB}
        \mathcal{L}_{disc}\left( \boldsymbol{\sigma}\right)
        \leq \sum_{t=T}^{1}\lambda_{t}\left\lVert D_{\theta} \left( \boldsymbol{\hat{x}}_t,\sigma_t \right) -\boldsymbol{x} \right\rVert^{2} +2\sigma_{1}^2,
\end{equation}
according to Jensen Inequality. Since the minimum noise level $\sigma_T$ is set as hyper-parameter, we simply ignore it here. Eq.~\ref{UB} shows the discretization loss in Eq.~\ref{loss} is upper-bounded by a diffusion loss with weighting factors $\lambda_i=\frac{\sigma_T}{\sigma_{i+1}}-\frac{\sigma_T}{\sigma_i}$. It can be efficiently optimized \wrt model $D_\theta$ in the same way as Eq.~\ref{diffusion_loss}.
Put it all together, we propose a two-stage finetuning method for pretrained diffusion models given $T$ steps sampling budget:
\begin{enumerate}
    \item In stage-one, learn the optimal sampling schedule $\boldsymbol{\sigma}$ for current denoising model $D_\theta$ by minimizing Eq.~\ref{loss};
    \item In stage-two, finetune the denoising model $D_\theta$ to current $\boldsymbol{\sigma}$ by minimizing the upper-bound Eq.~\ref{UB}.
\end{enumerate}

\subsection{Implementation}
For a $T$-step sampling schedule, we set the maximum and minimum noise levels to those supported by the pretrinaed models, and use a learning variable with $T-2$ degrees of freedom to model the increments following\cite{watson2021learning}. This is to ensure the sampling schedule is a valid monotonic function. Specifically, for noise level $\sigma_i$ it is parameterized as:
\begin{equation}
\begin{split}
\label{sigma_model}
        \sigma_t=\sum_{j=T}^{t}Softmax(\Bar{v}_j),
\end{split}
\end{equation}
where $\Bar{\boldsymbol{v}}=Concat\left(\boldsymbol{v};1\right)$ and $\boldsymbol{v}$ is the learning tensor.

Directly optimizing Eq.~\ref{loss} needs to back-propagate loss through all steps, memory consumption grows linearly with number of steps $T$ as the denoising model is called each iteration. Watson~\etal~\cite{watson2021learning} applies gradient rematerialization for back-propagating per-sample loss across diffusion generation process. They store all the $\left\{\boldsymbol{\hat{x}}_t\right\}$ sequence in memory and perform an extra forward pass of the corresponding inputs to re-build the computational graph at each step. In this case time consumption grows linearly with $T$. In our structure however, the discretization loss Eq.~\ref{loss} can be efficiently optimized without extra time or memory cost. Recall in the first equation of Eq.~\ref{preUB} discretization loss is represented as a function of $\left(D_\theta \left( \boldsymbol{\hat{x}}_i, \sigma_i\right)-\boldsymbol{x}\right)$. Gradient of Eq.~\ref{loss} \wrt each $\sigma_i$ can be written as:
\begin{equation}
\label{BPTD}
    \begin{split}
        \frac{\partial \mathcal{L}_{disc}\left( \boldsymbol{\sigma}\right)}{\partial \sigma_t}=2\lambda_t\left\langle\boldsymbol{\hat{x}}_1-\boldsymbol{x},\frac{\partial \left(D_\theta \left( \boldsymbol{\hat{x}}_t, \sigma_t\right)-\boldsymbol{x}\right)}{\partial \sigma_t}\right\rangle,
    \end{split}
\end{equation}
where the derivative term is available as a by-product of back-propagting diffusion loss $\left\lVert D_{\theta} \left( \boldsymbol{\hat{x}}_t,\sigma_t \right) -\boldsymbol{x} \right\rVert^{2}$. In practice, we jointly optimize Eq.~\ref{loss} and the diffusion loss at each step, with a loss scaler $\gamma$ as hyper-parameter. The overall framework of our method is listed in Algo.~\ref{alg}.
\section{Experiment}
\label{experiment}

In this section, we conduct several finetuning experiments using our two-stage method specified in Algo.~\ref{alg}. We first investigate the finetuning improvements of our method across pretrinaed models with different parameterization to various sampling steps in \S.~\ref{improve}. We then conduct ablation experiments on the challenging ImageNet dataset\cite{deng2009imagenet} to examine the effects of individual components in \S.~\ref{ablation}. Finally, based on the experiment results and our theoretical analysis in \S.~\ref{method}, we delve into a detailed discussion of the deep connection between diffusion generation process and weighting scheme of training objective in \S.~\ref{analysis}. We employ Fr\'echet inception distance (FID)\cite{heusel2017gans} in terms of sample quality assessment, as it is the most widely applied indicator in the literature and the results of different models are available for direct comparison. The FID results reported in this section is computed between 50K generated images and the training set following\cite{karras2022elucidating}.

\subsection{Finetune to Sampling Budgets}
\label{improve}
\begin{figure*}
  \centering
    \includegraphics[width=1.0\linewidth]{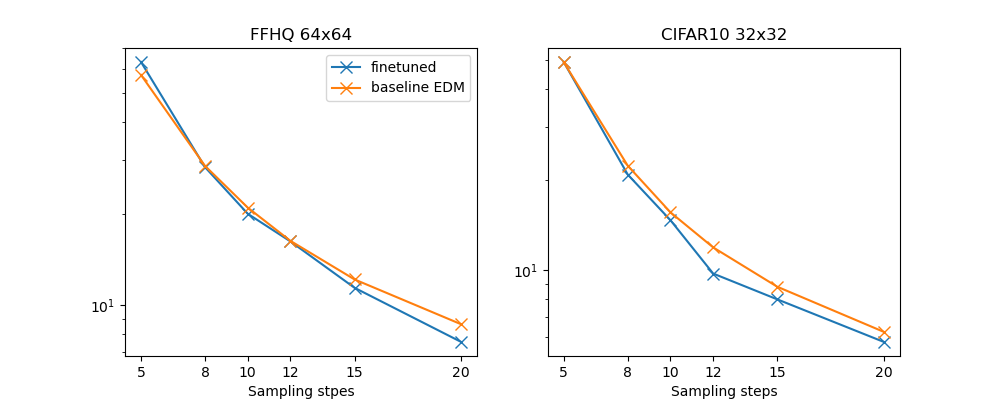}
  \caption{\textbf{Finetune improvement.} (Left) finetuning experiments conducted on pretrained FFHQ model at 64$\times$64 resolotion and CIFAR-10 model at 32$\times$32 resolotion (Right). Our finetuning can achieive consistent improvements on few-steps sampling.}
  \label{fig:result}
\end{figure*}
We finetune the pretrained Elucidated Diffusion Model (EDM)~\cite{karras2022elucidating} to given sampling steps, swapping from $T=\left\{5,8,10,12,15,20\right\}$. We adopt a model trained on downsampled FFHQ dataset at resolution 64$\times$64, and a model trained on CIFAR-10 dataset at resolution 32$\times$32. The models are finetuned for 50K parameters update using Algo.~\ref{alg}, and the finetuning improvements is presented in Fig.~\ref{fig:result}. On both datasets, our method yield approximate an improvement of 1 in FID score across different sampling budgets. The maximal improvement is observed at 20-steps finetuning on FFHQ model, while on CIFAR-10 model the maximum is observed at 12-steps finetuning. We suggest that this is because of the diversity of FFHQ dataset, which implies a more complicate ODE function for learning.

\subsection{Ablation Study}
\label{ablation}
We conduct the ablative experiments on the challenging ImageNet dataset\cite{deng2009imagenet} to examine the effectiveness of individual components in Algo.~\ref{alg}. The dataset are also pre-processed to 64$\times$64 resolution following\cite{karras2022elucidating}. We also perfomr 50K model parameters update on each of the configuration and present the result in Table.~\ref{abl_table}. The experiment results show that both learning the sampling schedule in stage-1 and the model finetuning in stage-2 is sufficient for improvements. We also finetune the pretrained model with original weighting scheme~\cite{karras2022elucidating} to testify our training objective induced by the learned sampling schedule. The result show finetuning following the original weights leads to a deterioration in model performance. This is because the original objective is derived in the continuous diffusion context which cannot be directly transfered to the discrete diffusion training. We provide further discussion on the connection between discrete sampling schedule and training objective in next section.

\begin{table}
  \caption{FID score on ImageNet64}
  \label{abl_table}
  \centering
  \begin{tabular}{lcc}
    \toprule
    {\bf Sampling Steps}    & {$T=20$} & {$T=10$}\\
    \midrule
    baseline (EDM)    & 6.448  & 22.42 \\
    w/ stage-1  &  6.138 & 14.86 \\
    w/ stage-2  &  6.293 & 16.89 \\
    original weight finetune & 6.537 & 25.17 \\
    two-stage  & \textbf{5.824} & \textbf{12.28} \\
    \bottomrule
  \end{tabular}
\end{table}

\subsection{Analysis}
\label{analysis}
\paragraph{Learned sampling schedule}
\begin{figure}[t]
  \centering
   \includegraphics[width=1.0\linewidth]{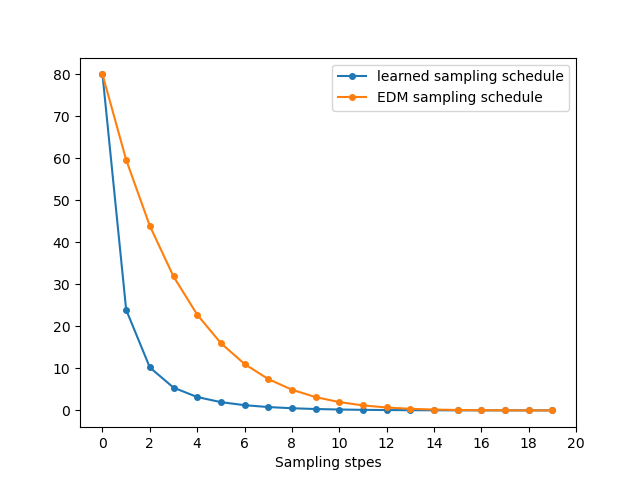}

   \caption{\textbf{Sampling schedule}. The learned sampling schedule of our method drastically reduces the noise level, skipping sampling steps where little content is generated and allocating computational resources to steps that lead to diverse perceptual details.}
   \label{fig:sigmas}
\end{figure}
In the baseline pretrianed models of Karras~\etal~\cite{karras2022elucidating}, they employ a convex function to provide the sampling schedule with a hyperparameter $\rho$ controlling the curvature. They found in experiments that using a sampling schedule with moderate curvature that smoothly decreases the noise levels leads to favorable results, which is illustrated by the orange line in Fig.~\ref{fig:sigmas}. Our two-stage finetuning method can lead to a sampling schedule with larger curvature, and we denote it by the blue line in Fig.~\ref{fig:sigmas}. The learned sampling schedule aggressively skips the early sampling steps with large noise levels, where previous work\cite{song2020denoising,nichol2021improved} found little contribution to the generation quality. However, these methods empirically choose sampling schedules according to the observation, while our method can learn this trend by directly optimizing the diffusion generation process. In EDM, although the curvature of the sampling schedule can directly controlled by the hyperparameter $\rho$, they found a too large $\rho$,\ie, skipping too much early steps is detrimental to generation quality. We suggest that this is attributed to the finetuning of the pretrained models in stage-2 of our method, results in smoother denoise outputs trajectory $\left\{\boldsymbol{\tau}_{\sigma^*_i}\right\}$ \wrt the learned sampling schedule as proposed in Theo.~\ref{theorem}. This phenomenon is more pronounced when the sampling budget is further limited, where the sampling schedule drastically reduce noise levels, leaving as much computational resources to lower noise level as possible.
\paragraph{Learned weighting scheme}
\begin{figure*}
  \centering
    \includegraphics[width=1.0\linewidth]{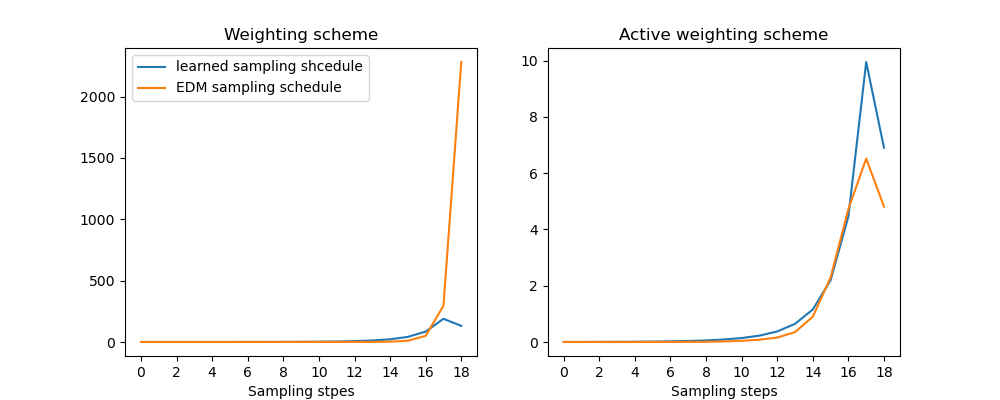}
  \caption{\textbf{Weighting schemes.} (Left) the explicit weighting scheme $\lambda_i$ in training objective Eq.~\cref{diffusion_loss,loss}. (Right) the active weighting scheme by including the likelihood term in Monte-Carlo estimating Eq.~\cref{diffusion_loss,loss}. The weighting scheme induced by our learned sampling schedule exhibits a same pattern as the empirically chosen one in EDM\cite{karras2022elucidating} which leads to better results in experiments.}
  \label{fig:weights}
\end{figure*}
The weighting scheme of diffusion training objective influences model performance to a large extent. In our structure, the weighting scheme is determined by learned sampling scheduled, while in previous work it is either derived from different theoretical perspective\cite{sohl2015deep,song2019generative,kingma2021variational,karras2022elucidating} or chosen empirically\cite{ho2020denoising,dhariwal2021diffusion,choi2022perception}. We compare the weighting scheme corresponds to our learned sampling schedule, to that employed by training EDM\cite{karras2022elucidating} in Fig~\ref{fig:weights}. We scale the weights of EDM training objective by $10^{-1}$ to directly compare two weighting schemes in one graph. Note that this corresponds to applying a loss-scaler during training. Our weighting scheme exhibits a sharp increase at the second last step, while the weighting function of EDM appears smoother. This is because EDM training objective is derived from the theoretical continuous diffusion model, the weighting scheme is also a continuous increasing function over time. However, the noise level where a continuous diffusion model is trained on follows a continuous probability distribution, which corresponds to a likelihood term in the Monte Carlo estimating loss function. EDM employted a Gaussian distributed noise level whose mean and variance are controlled by hyperparemeters. In our method as we finetune the models to a a given number of sampling steps, we sample noise levels with equal probabilities following\cite{kingma2021variational}. By explicitly including the likelihood terms, we plot the active weighting scheme in Fig.~\ref{fig:weights} (Right). The active weights of EDM continuously increase before reaching a threshold near the minimum noise level. They argue that models benefits little from training on noise levels that are either too large or too small. In a too large noise level with substantial information loss, the reconstruction result will approach dataset mean; in a too small noise level, the reconstruction loss can hardly provide any signal for updating model parameters. This is actually aligned with previous work observation\cite{dhariwal2021diffusion}. The weighting scheme induced by our learned sampling schedule exhibits the same pattern with these empirical findings. On the other hand, the expansion of Eq.~\ref{expansion} implies that the iterations toward the later steps will have higher weights in the last denoised output, \ie, have a direct impact on generation quality.
\section{Related Work}

\subsection{Diffusion Generative Modelsx+x}
Diffusion model is a class of generative models first proposed by Sohl-Dickstein~\etal~\cite{sohl2015deep} in the spirit of modeling the gradual changes in the distribution of a molecular system. Diffusion model generates images from noise by iteratively sampling from conditional distributions. Ho~\etal~\cite{ho2020denoising} significantly improved the performance by introducing a large U-Net model\cite{ronneberger2015u} to model this denoise process, demonstrating the potential of such models. In a concurrent work, Song~\etal~\cite{song2019generative} proposed Score-Based Generative Models which employs Langevin Dynamics to model the continuous changes in the data distribution. Although these two models are proposed from different theoretical frameworks, they bear a strong resemblance in both the inference process and training objectives, and were later unified into a generalized model by Song~\etal~\cite{song2020score}. Dhariwal~\etal~\cite{dhariwal2021diffusion} reported for the first time diffusion-based models outperform generative adversarial models\cite{goodfellow2020generative,karras2020training,sauer2022stylegan} on various tasks by further increasing model size. Besides, they applied an additional classifier to provide category information for effective conditional generation. Building upon these works, the influential work of Rombach~\etal~\cite{rombach2022high} proposed latent diffusion model. They applied a variational auto-encoder\cite{kingma2013auto,van2017neural} to project image data into latent space for diffusion process, addressing the previous limitation of diffusion model (LDM) in generating high-resolution images. Moreover, pretrained CLIP model\cite{radford2021learning} are used to provide text embeddings as additional input achieving text-to-image generation. Together, LDM brought generation quality and diversity to a new level.
\subsection{Diffusion Training Objective}
The original diffusion training objective was derived from the variational lower-bound (VLB) of data sample negative log-likelihood (NLL)\cite{sohl2015deep}. This objective can be simplified as a weighted summation of denoising loss at each step\cite{song2020score}. However, it was originally observed in experiments\cite{ho2020denoising,song2019generative} that a simple unweighted objective can somehow lead to better results. As the VLB objective theoretically guarantee better NLL, a combination of the simple loss and the VLB loss was also attempted\cite{dhariwal2021diffusion}. Essentially, this hybrid objective adopted a linear combination of two weighting schemes and empirically found improvements in both samples density estimation and FID score. Recent work\cite{kingma2021variational,song2021maximum} proposed some new weighting schemes for maximum likelihood training. Kingma~\etal~\cite{kingma2021variational} rewrite the objective as a function of signal-to-noise ratio (SNR), and found the original weights in VLB objective are actually the changing rate in SNR. By directly learning the SNR function , they achieved state-of-the-art density estimation.
\subsection{Diffusion Sampling}
Diffusion sampling process was constraint to the noise levels used during training\cite{ho2020denoising}. The generalized class of continuous diffusion models can perform sampling steps on arbitrary noise level within range supported by the models\cite{song2020score}, providing possibilities for exploring diffusion sampling algorithms. Previous work\cite{nichol2021improved,dhariwal2021diffusion} empirically found that the majority of perceptual content is generated when the noise level is low, and faster sampling can be achieved by reducing sampling steps with large noise levels. However, these methods still requires multiple iterations at lower noise levels. Song~\etal~\cite{song2020denoising} proposed a deterministic sampler named DDIM which can yield comparable results with 10 times faster generation speed. They manipulated the underlying probabilistic graphical and skipped a substantial number of iteration steps. From the theoretical perspective\cite{song2020score}, this method corresponds to  taking large step in numerical simulating as we analyzed in the main body of this paper.
\section{Conclusion}

In this study, we explore the possibility of finetune pretrained diffusion models to optimal ODE solver for a given iteration steps. We provide a detailed analysis on the truncation error made by common first-order diffusion sampler, find a simple formulation of the truncation error that represented by a function of denoising loss at each step. We propose a efficient method to directly optimize diffusion generation process without extra time or memory cost. The theoretical analysis of truncation error shows that the optimal sampling schedule can lead to a first order global error. In this light we propose a two-stage finetuning method that learn the optimal sampling schedule in stage-1 and finetune the pretrained model according to the learned schedule in stage-2. We hope this novel methodology can inspire future explorations towards better sampling algorithms for diffusion models.

{\small
\bibliographystyle{ieee_fullname}
\bibliography{reference}
}

\end{document}